\documentclass[10pt,twocolumn,letterpaper]{article}

\usepackage[pagenumbers]{iccv}

\definecolor{iccvblue}{rgb}{0.21,0.49,0.74}
\definecolor{urlred}{rgb}{1,0.2,0.6}
\usepackage[pagebackref,breaklinks,colorlinks,allcolors=iccvblue,urlcolor=urlred]{hyperref}
\usepackage{bbm}
\usepackage{multirow}
\usepackage{tabularx}
\usepackage{algorithm}
\usepackage{algorithmic}
\usepackage{colortbl}
\usepackage{makecell}

\usepackage{marvosym}

\usepackage[accsupp]{axessibility}  

\definecolor{yellow}{rgb}{1, 1, 0.7}
\definecolor{orange}{rgb}{1, 0.85, 0.7}
\definecolor{tablered}{rgb}{1, 0.7, 0.7}

\def\ourmodel{DriveX}

\def\confName{ICCV}
\def\confYear{2025}

\title{Driving View Synthesis on Free-form Trajectories with Generative Prior}

\author{Zeyu Yang$^1$\footnotemark[1] \quad Zijie Pan$^1$\footnotemark[1] \quad Yuankun Yang$^1$\footnotemark[1]\quad Xiatian Zhu$^2$\quad Li Zhang$^1$\footnotemark[2]
\\
$^{1}$ Fudan University \qquad $^{2} $ University of Surrey
\vspace{.5em} 
\\
\href{https://fudan-zvg.github.io/DriveX}{\texttt{fudan-zvg.github.io/DriveX}}
} 

\begin{document}
\maketitle

\begin{figure*}[t]
\centering
\includegraphics[width=\linewidth]{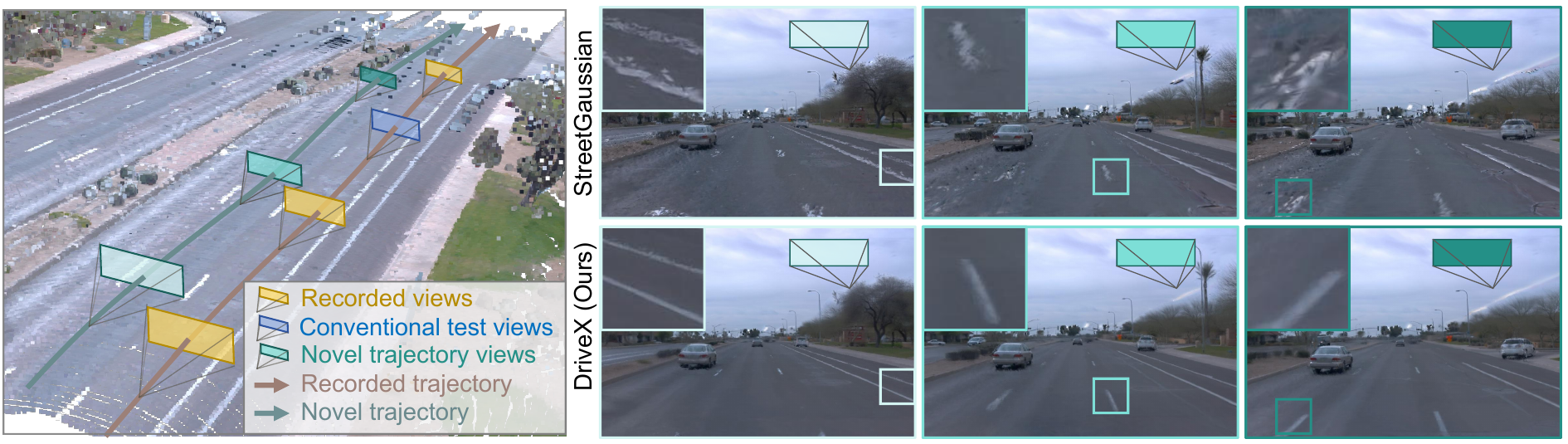}
\vspace{-7mm}
\caption{
Though conventional reconstruction-based approaches can well fit views on the recorded trajectory, they struggle to extrapolate onto the views beyond it (\textbf{top}). 
In contrast, our method significantly improves the synthesis quality on these novel trajectory views by incorporating generative prior (\textbf{bottom}), thereby achieving superior synthesis quality on free-form trajectories.
}
\label{fig:teaser}
\vspace{-6mm}
\end{figure*} 
\renewcommand{\thefootnote}{\fnsymbol{footnote}}
\footnotetext[1]{Equally contributed.}
\footnotetext[2]{Li Zhang (\href{mailto:lizhangfd@fudan.edu.cn}{\texttt{lizhangfd@fudan.edu.cn}}) is the corresponding author with School of Data Science, Fudan University.}

\begin{abstract}
Driving view synthesis along free-form trajectories is essential for realistic driving simulations, enabling closed-loop evaluation of end-to-end driving policies. Existing methods excel at view interpolation along recorded paths but struggle to generalize to novel trajectories due to limited viewpoints in driving videos. 
To tackle this challenge, we propose \textbf{\ourmodel{}}, a novel free-form driving view synthesis framework, that progressively distills generative prior into the 3D Gaussian model during its optimization.
Within this framework, we utilize a video diffusion model to refine the degraded novel trajectory renderings from the in-training Gaussian model, while the restored videos in turn serve as additional supervision for optimizing the 3D Gaussian.
Concretely, we craft an inpainting-based video restoration task, which can disentangle the identification of degraded regions from the generative capability of the diffusion model and remove the need of simulating specific degraded pattern in the training of the diffusion model.
To further enhance the consistency and fidelity of generated contents, the pseudo ground truth is progressively updated with gradually improved novel trajectory rendering, allowing both components to co-adapt and reinforce each other while minimizing the disruption on the optimization.
By tightly integrating 3D scene representation with generative prior, \ourmodel{} achieves high-quality view synthesis beyond recorded trajectories in real time--unlocking new possibilities for flexible and realistic driving simulations on free-form trajectories.

\end{abstract} \section{Introduction}
\label{sec:intro}

Building virtual driving worlds capable of being engaged with driving policies plays a pivotal role in developing robust autonomous driving systems. It facilitates efficient training data scaling and automatic synthesis of safety-critical long-tail cases, while enabling close-loop evaluation for end-to-end autonomous driving systems.

In recent years, the burgeoning advancements in novel view synthesis~\cite{mildenhall2020nerf,kerbl20233d} have propelled the prevalence of reconstruction-based driving simulations~\cite{xie2023s,chen2023periodic,yan2024street}.
These methods aim to reconstruct driving environments from vehicle sensor data typically involving single-trajectory videos from surrounding cameras and synchronized LiDAR point clouds.
However, the limited overlap between frames, large textureless regions and illumination variations in large-scale scenes, drastically complicate this task than conventional novel view synthesis.
Consequently, despite achieving impressive results in fitting training views and interpolating along recorded trajectories, they still struggle to extrapolate on novel views far from the recorded trajectory, limiting their flexibility and usefulness in many downstream tasks.

Recent works~\cite{liu2024reconx,liu20243dgs,yu2024viewcrafter} have demonstrated that 2D generative priors can effectively regularize sparse-view reconstruction by synthesizing pseudo ground-truth images for novel view, thereby alleviating the under-constraint nature of this task. 
However, directly applying these approaches in the driving scenes presents unique challenges due to the limited diversity of trajectories in existing driving video datasets. The training is primarily confined on forward-moving trajectories, while the off-trajectory videos are required for inference. 
This will introduce a considerable gap, making it infeasible to learn precise explicit pose control, and posing a great challenge to simulate degraded patterns for training a restoration-based model.

To address these challenges, we propose a reconstruction-generation interwined framework for driving view synthesis along free-form trajectories, termed \textbf{\ourmodel{}}, which progressively distills video generative prior into the 3D scene representation (\ie, 3D Gaussian Splatting~\cite{kerbl20233d}). During this process, a video diffusion model is employed to generate novel trajectory videos as additional supervision for the optimization of scene representation, effectively mitigating the sparse-view issue. 
This is achieved by leveraging a generative model to restore novel trajectory renderings from the in-training 3D Gaussian model, which ensures the consistency of generated content while eliminating the need for explicit pose condition.
Specifically, we design an inpainting-based restoration task where generation is conditioned on the rendered video masked by a geometry-aware unreliability mask. 
This formulation allows the model to fully leverage its generative capability, freeing it from the task of identifying degraded regions, thus circumventing the complexity of simulating degraded patterns during its training.
During the optimization of scene representation, we progressively perform the restoration process and update the generated pseudo ground truth to further enhance consistency of the generated content. As shown in~\cref{fig:teaser}, our approach substantially improves the view synthesis quality in the novel trajectory compared to the state-of-the-art reconstruction-only method~\cite{yan2024street}.

Our contributions are as follows:
\textbf{(i)} We propose a novel driving view synthesis framework that seamlessly integrates generative prior into driving scene reconstruction, where a diffusion model is employed to progressively synthesize consistent novel view ground truth for regularizing the optimization of the 3D Gaussian model;
\textbf{(ii)} By formulating an inpainting-based restoration task, we disentangle the identification of degraded regions from the generative capability of the diffusion model, thereby simplifying training on driving videos with limited diverse trajectories;
\textbf{(iii)} Our DriveX effectively overcomes the sparse-view issue in previous street scene reconstruction methods, enabling high-fidelity view synthesis along free-form trajectories.

 \section{Related works}
\label{sec:related}

\noindent{\bf Reconstruction-based driving view synthesis}~~
Early approaches~\cite{xie2023s, yang2023emernerf,tancik2022block, wang2023neural,turki2023suds} for dynamic urban scenes reconstruction focus on extending neural radiance fields (NeRF)~\cite{mildenhall2020nerf} in large scale unbounded scenes. These methods are restricted with low efficiency and blurry rendering performance. 
Recently, another line of works introduced 3D Gaussian splatting~\cite{kerbl20233d} into this task to produce real-time high-fidelity rendering. Some of them~\cite{yan2024street,zhou2024drivinggaussian} leverage its favorable explicit property to decouple the whole scene into the static background and moving foreground Gaussians, while others modeling the motion of 3D Gaussians with periodic functions~\cite{chen2023periodic} or deformation fields~\cite{huang2024s3gaussian}.
However, driving scenes featured with sparse views, little overlaps, and large textureless areas,
making existing methods struggle to generalize to substantively novel views.

\noindent{\bf Generative-based driving view synthesis}~~
Recently, propelled by the significant advancement of the diffusion model~\cite{ho2020denoising,song2020score,rombach2022high,song2023consistency,blattmann2023stable}, many works employ video generators to construct world models~\cite{lecun2022path}.
Some of them introduce this paradigm into autonomous driving~\cite{wang2023drivedreamer,zhao2024drivedreamer,li2023drivingdiffusion,gao2023magicdrive,wang2024driving,gao2024vista,hu2023gaia, lu2025wovogen} for 
simulating the plausible visual outcomes under versatile driving controls.
Despite the remarkable diversity of generated results, this approach is constrained by the absence of underlying 3D model.
Consequently, they cannot promise geometry and texture consistency for the generated many-trajectory videos of the same scene.

\begin{figure*}[t]
    \centering
    \includegraphics[width=\textwidth]{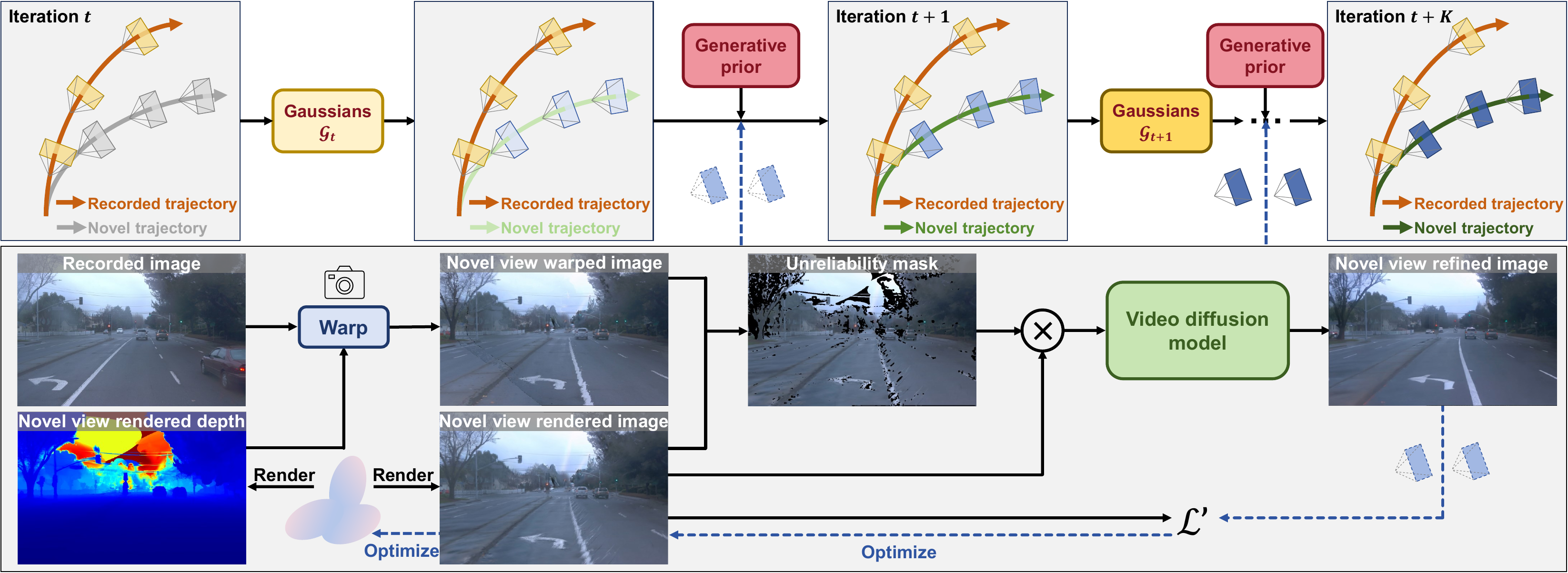}
    \vspace{-6.5mm}
    \caption{
    \textbf{Schematic illustration of \ourmodel}. 
    \textbf{Top: Optimizing the scene representation} with both recorded trajectory views and generated novel trajectory views as supervision. \textbf{Bottom: Generative prior integration.} During optimization, given a novel view rendering from in-training Gaussian $\mathcal{G}_t$, a video diffusion model is employed to generate a restored version as pseudo ground truth on novel trajectories, guided by its unreliability mask and accumulated LiDAR projections. 
    }
    \label{fig:pipeline1}
    \vspace{-5.5mm}
\end{figure*} 
\noindent{\bf Scene reconstruction with diffusion prior}~~
Leveraging generative prior to enhance the reconstruction from sparse view observations~\cite{yu2024viewcrafter,liu20243dgs,liu2024reconx,wu2024reconfusion} has emerged.
Extending this paradigm into driving scene reconstruction has been recently attempted \cite{yu2024sgd,hwang2024vegs,zhao2024drive}. 
Limited diversity of trajectory in driving scene data poses unique challanges for them.
SGD~\cite{yu2024sgd} and VEGS~\cite{hwang2024vegs} contribute to define and tackle pseudo and extrapolated view synthesis in driving scenarios. 
However, these methods may encounter challenges with accurate pose and detail estimation under novel trajectories. 
DriveDreamer4D~\cite{zhao2024drive} samples novel trajectory video based on single reference frame before training, risking hallucination and inconsistency due to insufficient constraints. StreetCrafter~\cite{yan2024streetcrafter} refines the novel view renderings using a controllable video diffusion model guided by LiDAR projections. 
Meanwhile, FreeSim~\cite{fan2024freesim} and Difix3D+~\cite{wu2025difix3d} directly condition on degraded renderings and employ carefully data curation to train the generative model to adapt on degradation patterns resulted from off-trajectory renderings. 
In contrast, by crafting an inpainting-based video restoration task, we free the generative model from identifying degraded regions, thereby eliminating the need to simulate specific degraded patterns in training.

\section{Methodology}
\label{sec:method}

\subsection{Preliminary: Driving scene reconstruction
}
\label{sec:preliminary}
Reconstruction-based driving simulation aims to recover the surrounding scenes from calibrated video 
$\mathcal{V}_{\mathrm{gt}}=\left\{ I_{i} \right\}_{i=0}^F$ recorded along a driving trajectory $\mathcal{T}=\left\{P_{i}\right\}_{i=0}^F$, where $P_{i} \in \mathrm{SE(3)}$ is the camera pose at frame $i$.
Then the sensor data can be simulated under the unrecorded actions ({\em e.g.}~lane change) in the reconstructed environment.

Recently, due to the superior fidelity and efficiency of 3D Gaussian splatting (3DGS)~\cite{kerbl20233d}, it has been widely employed in numerous street scene reconstruction efforts~\cite{yan2024street,zhou2024drivinggaussian,chen2023periodic,huang2024s3gaussian} as a fundamental scene representation. 
3DGS represents scenes as a set of 3D Gaussians, denoted $\mathcal{G}$. The influences of each Gaussian primitive with position $\mathbf{x}_k \in \mathbb{R}^{3}$ and covariance $\mathbf{\Sigma} \in \mathbb{R}^{3 \times 3}$ on any position is given by an unnormalized Gaussian function $G_k(\cdot;\mathbf{x}_k,\mathbf{\Sigma})$ weighted by its opacity $o_k \in\mathbb{R}$.
Given \textit{any} camera pose $P_i$, it can be rendered by performing splatting and alpha blending on $N$ sorted Gaussians visible at this view:
\begin{equation}
    I^{\prime} = R(\mathcal{G}, P_{i}) =
    \sum^{N}_{k=1}  o_k G_{k}  c_k(P_{i})
    \prod^{k-1}_{j=1} (1- o_j G_{j}),
    \label{eq:3DGS_rendering}
\end{equation}
where $R$ represents the differentiable Gaussian rasterizer, $I^{\prime}$ denotes the rendered image, $c_k$ is the view-dependent color of the $k$-th Gaussian.
Then the Gaussians $\mathcal{G}$ can be optimized via the photometric loss between rendered images $I_{i}^{\prime}$ and ground truth $I_{i}$ from recorded video $\mathcal{V}_{\mathrm{gt}}$ by:
\begin{equation}
    \label{eq:loss}
    \mathcal{L}_{\text{img}}(I_{i}^{\prime}, I_{i}) = \lambda \Vert I_{i}^{\prime} - I_{i}\Vert_1 + (1-\lambda) \mathcal{L}_\text{SSIM}(I_{i}^{\prime}, I_{i}),
\end{equation}
where $\mathcal{L}_\text{SSIM}$ is the SSIM~\citep{wang2004ssim} loss and $\lambda$ is a weight.

Following~\cite{yan2024street}, we model the traffic participants in the driving scenes as a dynamic node in the neural scene graph with rigid motion, which can be regarded as static in their own local coordinate frames.
The transformation from each object's local frame to the global coordinate frame of the whole scene is parameterized by a series of rotations and transformations, which can be annotated by their tracked bounding boxes. 
Additionally, cube map is employed to model the sky with infinite distance~\cite{yan2024street,chen2023periodic}.

\noindent{\bf Challenges}~~
Since the camera configuration in autonomous vehicles is primarily designed for perception tasks, the captured videos often offer limited viewpoints, minimal overlap, and large textureless regions.
These inherent properties bring challenges to the reconstruction of the unbounded driving scenes from single-trajectory videos.
Fitting the training views at recorded trajectory $\mathcal{T}$ is insufficient to realize satisfactory extrapolation out of $\mathcal{T}$ for novel view synthesis.
As a result, existing optimization-based driving scene synthesis approaches struggle to achieve high-quality rendering along novel trajectories $\mathcal{T}^{\prime}=\{P^{\prime}_i|P^{\prime}_i \notin \mathcal{T}\}_{i=0}^F$, often leading to obvious degeneration and artifacts.

\subsection{Novel view regularization with generative prior}
\label{sec:3.2}
\noindent \textbf{Generate consistent novel view via video restoration}
A promising solution to the aforementioned challenges is to incorporate generative priors as an additional regularization. Large video generative models trained on internet-scale data have developed a strong knowledge of natural video distributions, enabling them to reliably infer and generate novel view content as pseudo ground truth for optimizing 3DGS. 
To ensure consistency between the generated content and the underlying scene, thereby preventing disruption to optimization, the generation should be appropriately conditioned on the in-training Gaussian model.

To achieve this, we delicately design a video restoration task, 
which is formulated as recovering clean novel view images $\mathcal{V}$ from the artifact-heavy novel view renderings $\mathcal{V}^{\prime}$ beyond the recorded trajectory. Within this design, a video generative model can be employed to address this problem effectively. The refined video then serves as additional supervision in the optimization of 3DGS.

Concretely, given a sequence of viewpoints $\mathcal{T}^{\prime}=\left\{P_{i}^{\prime}\right\}_{i=0}^F$ along a novel trajectory,
a video diffusion model $\mathcal{D}$ is employed to estimate clean version $\mathcal{V}_i$:
\begin{align}
\label{eq:render}
\mathcal{V}^{\prime}_t &= \left\{ R(\mathcal{G}_{t}, ~ P_{i}^{\prime}) \right\}_{i=0}^F, \\
\label{eq:refine_ori}
\mathcal{V}^{\prime}_{t, \mathrm{refine}} &= \mathcal{D} \left( \mathcal{V}^{\prime}_t \right), \\
\label{eq:novel_loss}
\mathcal{L}^{\prime} &\triangleq \mathcal{L}_{\text{img}}(\mathcal{V}^{\prime}_t, ~ \mathcal{V}^{\prime}_{t,\mathrm{refine}}),
\end{align}
where $\mathcal{V}^{\prime}_t$ is rendered from the Gaussian model $\mathcal{G}_{t}$ at $t$-th iteration of optimization through \cref{eq:3DGS_rendering}, the refined video $\mathcal{V}^{\prime}_{t, \mathrm{refine}}$ can be regarded as the estimation of $\mathcal{V}$. Typically, $\mathcal{D}$ takes noise-perturbed images as input, conditioned on the degraded video, and outputs the refined video $\mathcal{V}^{\prime}_{t, \mathrm{refine}}$. 
Then $\mathcal{V}^{\prime}_{t, \mathrm{refine}}$ is used as the pseudo ground truth in the novel trajectory loss $\mathcal{L}^\prime$ (\cref{eq:novel_loss}) to supervise the novel view renderings as illustrated in \cref{fig:pipeline1}.

\begin{algorithm}[t]
    \caption{Iterative refinement}
    \label{alg:iterative}
    \begin{algorithmic}[1]
    \REQUIRE Initial Gaussian model $\mathcal{G}_0$, novel trajectories $\mathcal{T}^{\prime}$, video diffusion $\mathcal{D}$, total steps $T$, warm up steps $T_0$
    \FOR{ $t = 0,\cdots,T-1$}
        \STATE $\mathcal{V}_t \leftarrow \left\{ R(\mathcal{G}_{t}, P_{i}) \right\}_{i=0}^F$
        \STATE Computing loss $\mathcal{L}_{\text{img}}(\mathcal{V}_t, \mathcal{V}_{\text{gt}})$
\IF{$t\ge T_0$}
            \STATE $\mathcal{V}^{\prime}_t \leftarrow \left\{ R(\mathcal{G}_{t}, P_{i}^{\prime}) \right\}_{i=0}^F$
            \IF{$(t-T_0)\mod K = 0$}
                \STATE $k \leftarrow t$
                \STATE $\mathcal{V}^{\prime}_{k, \mathrm{refine}} \leftarrow \mathcal{D} \left( \mathcal{V}^{\prime}_t, \mathcal{M} \right)$
            \ENDIF
            \STATE Computing loss $\mathcal{L}^\prime \leftarrow \mathcal{L}_{\text{img}}(\mathcal{V}^{\prime}_t, \mathcal{V}^{\prime}_{k,\mathrm{refine}})$
        \ENDIF
\STATE Backwarding loss and updating $\mathcal{G}_{t+1}$ 
    \ENDFOR
    \RETURN $\mathcal{G}_T$
    \end{algorithmic}
\end{algorithm}

\noindent \textbf{Iterative refinement} 
The generative model can be employed to provide novel view supervision for enhancing the novel view renderings. 
In turn, the generation quality also benefits from these improvements, as it is conditioned on $\mathcal{G}_t$ whose enhancements could provide more accurate information to guide the generation process.
This co-adapt and reinforced enhancement motivates an \textit{iterative refinement} in our framework. 
As outlined in \cref{alg:iterative}, we initiate the optimization for $T_0$ steps using conventional reconstruction-based method~\cite{yan2024street}. Subsequently, we perform video restoration in \cref{eq:refine_ori} based on the in-training Gaussians $\mathcal{G}_{T_0}$, and use a buffer to store refined videos for reuse in the following $K$ iterations. To ensure that the generated content can be improved along with the progressively optimized reconstruction results, we update all videos in the buffer every $K$ iterations by performing video restoration conditioned on the latest novel trajectory renderings.

Due to the additional time demands of running the video diffusion model, it is inefficient to set $K=1$, \ie, the buffer is updated in every iteration. We set it to $3000$ to strike the balance between quality and efficiency. Please refer to \cref{sec:exp_ablation} for the trade-off between different choice of $K$.
As the generated contents are conditioned on Gaussian model $\mathcal{G}_t$, they are constrained with the recorded scene, thus ensuring the consistency between the recorded and generated ground truth and leading to stable optimization.

\noindent \textbf{Novel trajectory sampling}
Another important design space is how to sample the novel trajectory $\mathcal{T}^{\prime}=\left\{P_{i}^{\prime}\right\}_{i=0}^F$ to be distilled by video diffusion model.
The ideal trajectory should balance the two requirements: (i) maximizing the utility of the generative model, \ie, the generated novel views should contribute as much as possible to enhance reconstruction quality; (ii) minimizing inconsistency between the recorded and generated images to mitigate potential perturbation for Gausssian's optimization. 
Therefore, we adopt a panning camera trajectory that starts from a recorded front view $P^{\prime}_0=[R_0|T_0]$ and gradually shifts laterally:
\begin{equation}
    P^{\prime}_i = [R_0 | \frac{i}{F} s \mathbf{v} + T_0],
\end{equation}
where $\mathbf{v} \in \mathbb{R}^3$ is the shifting direction, and $s\in \mathbb{R}$ controls the maximum shifting length. 
This design not only provides novel views beyond the recorded trajectory,
but also allows video generation model to extract detailed references from the initial frame.
At the early stage of optimization, rendering quality may degrade as the camera deviates from the recorded driving trajectory, which offers limited guidance for the video generation. To address this issue, we initially use a small shifting length $s$ and progressively extend it during optimization, ensuring our method’s stability and robustness across diverse scenes.

\subsection{Inpainting-based video restoration}
\label{sec:3.3}
Under the proposed trajectory sampling scheme with progressively increasing shifted length, degradations in the rendered video caused by the poor extrapolation capabilities of the scene representation are typically localized. 
Therefore, the key to effective video restoration lies in accurately identifying the artifacts in the rendered videos.
Ideally, a well-trained generative model should be capable of automatically identifying and restoring these degraded regions. 
However, limited viewpoints in driving video datasets make it challenging to simulate degradation caused by off-trajectory rendering while obtaining corresponding ground truth during the training of the generative model. 
Therefore, we propose an inpainting-based restoration framework that removes the need for the model to adapt to specific degraded patterns, thereby simplifying the training process.

Specifically, we construct an unreliability mask representing the artifacts via comparing the rendered images with the recorded images, 
which directly guides the model to focus on refining these unreliable regions while preserving reliable areas. Thus, the \cref{eq:refine_ori} should be revised as:
\begin{equation}
\label{eq:refine}
    \mathcal{V}^{\prime}_{t, \mathrm{refine}} = \mathcal{D} \left( \mathcal{V}^{\prime}_t, ~\mathcal{M} \right),
\end{equation}
where $\mathcal{M}$ is the mask indicating unreliable regions, and $\mathcal{D}$ is conditioned on the masked video.

\noindent \textbf{Unreliability mask} To assess geometry accuracy in novel view rendered images, we derive $\mathcal{M}$ by comparing the rendered images with novel view warped images from the recorded trajectory.
Let $I_{\text{ren}} \in \mathcal{V}^{\prime}_t$ denote a rendered image, and let $I_{\text{rec}}$ represent the recorded image that are ``closest'' to $I_{\text{ren}}$. 
We define a warping operation $\psi$ in \cref{eq:warp} to project a 3D point $p$ onto the image plane given camera pose $P$:
\begin{equation}
\label{eq:warp}
    (x, y, d) = \psi(p | P),
\end{equation}
where $x,y$ represent the pixel coordinates, and $d$ is the depth (Please refer to Appendix \textcolor{iccvblue}{B} for details). 
Note that the refined video starts from a front view on the recorded trajectory, which is selected as the ``closest'' image for all refined views in the video. 
Notably, $\psi$ is invertible, allowing us to unproject an image with its corresponding depth map back to 3D points given the camera pose. 
Consequently, we can obtain a pseudo image $\hat{I}_{\text{ren}}$ under the pose $P_{\text{ren}}$ by sampling the color in $I_{\text{rec}}$ using the warped pixel coordinates $(\mathbf{x}, \mathbf{y}, \mathbf{d})$ from unwarped $I_{\text{ren}}$ to $I_{\text{rec}}$, as expressed by \cref{eq:reprojection}.
\begin{equation}
\label{eq:reprojection}
    (\mathbf{x}, \mathbf{y}, \mathbf{d}) = \psi(\psi^{-1}(D_{\text{ren}} |P_{\text{ren}}) | P_{\text{rec}}),
\end{equation}
where $D_{\text{ren}}$ is the depth map corresponding to $\hat{I}_{\text{ren}}$ rendered by $\alpha$-blending each Gaussian center's depth.
Since it is hard to guarantee the pixel-wise correspondence between $I_{\text{ren}}$ and $\hat{I}_{\text{ren}}$, we choose SSIM to evaluate similarity at the patch-structure level.
The unreliability mask $\mathcal{M}$ is then obtained by applying a threshold $\tau$ to SSIM score:
\begin{equation}
\label{eq:mask}
    \mathcal{M} = \mathbbm{1}\left( \text{SSIM}(I_{\text{ren}}, \hat{I}_{\text{ren}}) < \tau \right).
\end{equation}
Because the warping process involves both rendered depth and rendered image, the discrepancy can serve as an indicator of geometric or appearance unreliability.

Since the warped image provides a rough overview of the novel trajectory, making it unnecessary to start denoising from scratch when we refine the videos using \cref{eq:refine}. Hence, we perturb warped videos by adding Gaussian noise at some noise level. 
This not only reduces unnecessary time costs but also avoids color shifts by preserving accurate low-frequency components in original video. 
Because autonomous vehicles are typically equipped with surround-view cameras, we include side view on recorded trajectory in the same frame as the auxiliary source images, which is also warped using equation~(\ref{eq:reprojection}) to the regions that are invisible by the front view to ensure well-defined unreliability masks for all areas of the novel view.

Since the sparse depth ground truth in the novel view can be derived from LiDAR measurements, the Gaussian model performs better in depth rendering than in color rendering under novel view, even in the presence of overfitting. 
This allows the warped images to serve as a reliable anchor for indicating the reliability.

\noindent \textbf{Projected LiDAR as auxiliary condition} Additionally, we also utilize colored LiDAR projections as an oracle to guide the generative process. For each novel view requiring restoration, LiDAR points projected onto it are accumulated from its adjacent $\pm 2$ frames, where points belonging to moving objects are aligned to the rendered frame based on their tracked poses. This empirically stabilizes the optimization, especially for those challenging scenes where rendered novel views struggle to provide reliable information in the early optimization. 
In light of the importance of novel view depth supervision within this framework, we additionally use the accumulated LiDAR point cloud as the depth ground truth for novel views.

\begin{figure*}[t]

\centering
\includegraphics[width=0.977\linewidth]{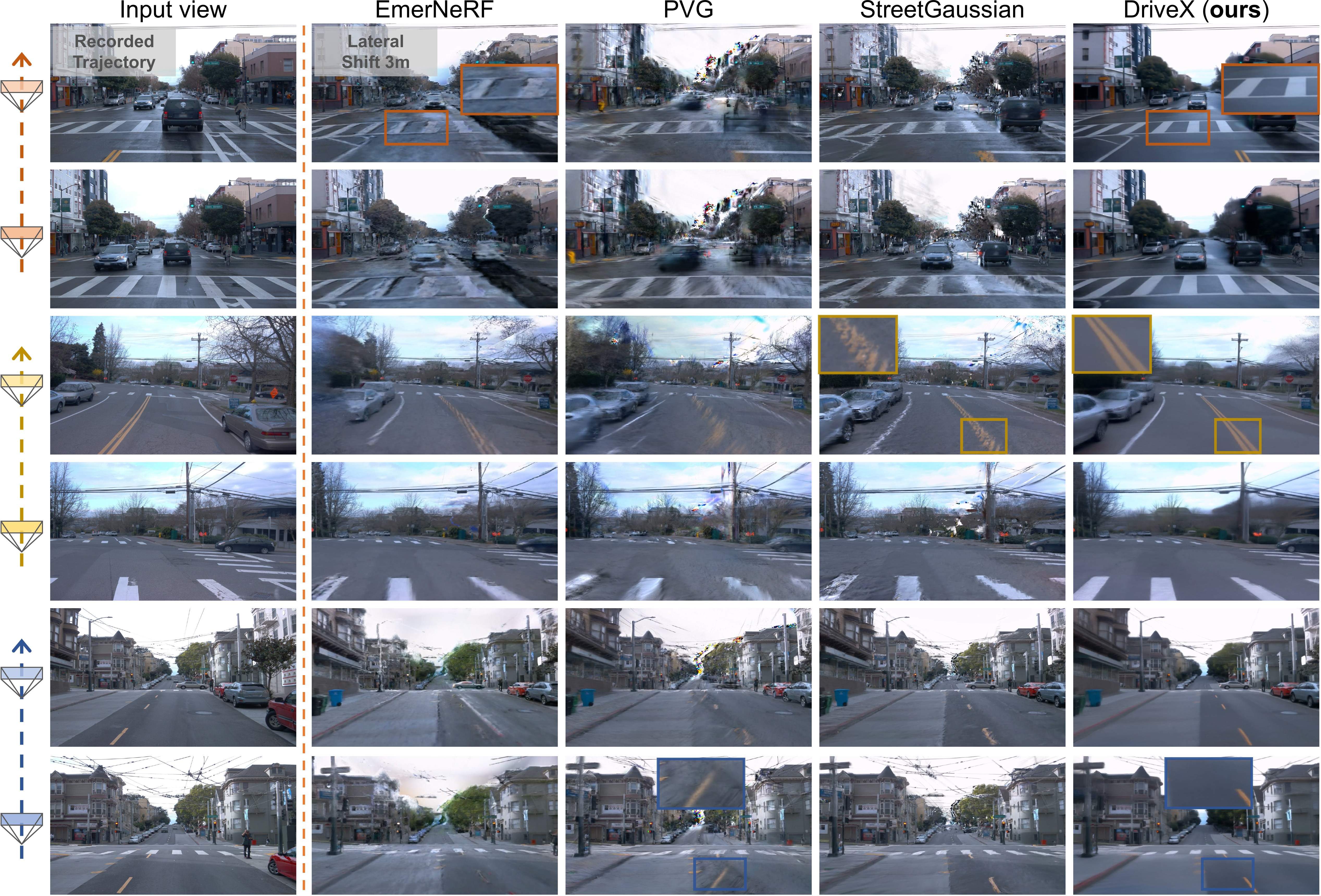}
\vspace{-2mm}
\caption{\textbf{Qualitative comparisons on novel trajectories.} 
For each method, images are rendered at the novel views with lateral shift of $3m$ to the recorded trajectories. 
}
\label{fig:qualitative}
\vspace{-5mm}
\end{figure*} 
\subsection{Training of generative model}
\label{sec:3.4}

One remaining question is how to obtain a generative prior capable of handling the restoration task in \cref{eq:refine}. 
Due to the limited trajectory in existing driving datasets, precisely simulating degradation in novel trajectory views caused by overfitting has an inherent challenge. Fortunately, our inpainting approach can mitigate this issue by disentangling the degraded pattern from the condition, avoiding the necessity to simulate degraded rendering during training and freeing the model from identifying potential degradations.

Specifically, we employ an edge-aware strategy to generate unreliability masks for the training data. This leads to a more general task which can cover scenarios typically encountered during off-trajectory rendering. 
Given a RGB image, we first convert it into grayscale. Then its directional derivatives computed using Sobel filters, as the edge intensity at each pixel. The resulting intensity map is normalized to yield a probability distribution over the pixels after adding a baseline offset for background regions. Then a predefined number of pixels is randomly sampled based on the distribution. For each sampled pixel, a $3\times3$ patch centered at the pixel is masked out to imitate the structured occlusion.
Both front and side-view camera videos are used in training. Because the trajectory of the side-view camera along the moving forward trajectory is similar to the lateral shift of the front cameras adopted in our novel trajectory sampling, as suggested by~\cite{wang2024freevs,fan2024freesim}.

While there remains a gap between training and inference, it does not substantially hinder the usage. This is because the training process primarily aims at unleashing the generative capability of the diffusion model to adapt on such a masked image inpainting task, rather than fitting specific mask patterns. 
Actually, even~\cite{yu2024viewcrafter} trained only on rendered point clouds can also serve as a competent generative prior for the crafted inpainting-based restoration task. 
While fine-tuning the model on driving videos can further adapt it to the style of driving scenarios, thereby significantly improving its robustness as demonstrated in~\cref{sec:exp_ablation}.

\begin{table*}[t]
\centering
    \setlength{\tabcolsep}{4pt}
\begin{tabular}{l|cc|ccc|ccc|ccc|c}\toprule
    & \multicolumn{2}{c|}{$\pm 0m$ (recorded)} & \multicolumn{3}{c|}{$\pm 1m$} & \multicolumn{3}{c|}{$\pm 2m$} & \multicolumn{3}{c|}{$\pm 3m$} & \multirow{2}*{FPS}
    \\
    & PSNR$\uparrow$ & SSIM$\uparrow$ & IoU$\uparrow$ & AP$\uparrow$ & FID$\downarrow$ & IoU$\uparrow$ & AP$\uparrow$ & FID$\downarrow$ & IoU$\uparrow$ & AP$\uparrow$ & FID$\downarrow$ &  \\
    \midrule
    \midrule
    EmerNeRF~\cite{yang2023emernerf} 
    & 29.60 & 0.8459
& 0.1147 & 0.5101 & 89.43
    & 0.0881 & 0.4644 & 102.98
    & 0.0735 & 0.4072 & 122.83
    & 0.12
    \\
    $S^3$Gaussian~\cite{huang2024s3gaussian}
    & 29.68 & 0.8670
& 0.0718 & 0.4499 & 115.67
    & 0.0401 & 0.4127 & 131.26
    & 0.0177 & 0.3812 & 162.03
    & 8.4
    \\
    PVG~\cite{chen2023periodic}  
    & \bf 29.98 & 0.8627
& 0.1082 & 0.5444 & 51.86
    & 0.0291 & 0.5102 & 86.93
    & 0.0170 & 0.4699 & 123.53
    & \textbf{48}
    \\
    StreetGaussian~\cite{yan2024street}  
    & 29.76 & \bf 0.8738
& 0.2671 & 0.5919 & 51.87
    & 0.2122 & 0.5848 & 70.66
    & 0.1720 & 0.5697 & 93.04
    & 34
    \\
\midrule
    \rowcolor[gray]{.9}
     \bf \ourmodel{} (ours) 
     & 29.74 & 0.8700
& \bf 0.2880 & \bf 0.5932 & \bf 46.91
   & \bf 0.2710 & \bf 0.5901 & \bf 65.52
   & \bf 0.2699 & \bf 0.5712 & \bf 79.78
     & 34
     \\
    \bottomrule
    \end{tabular}
    \vspace{-3mm}
    \caption{\textbf{Quantitative evaluation on Waymo dataset~\cite{sun2020scalability}.} 
    We compare images rendered in novel trajectories with different shift lengths, where the IoU and AP are computed between the projected ground truth and those detected on synthesized views by perception models~\cite{ren2015faster,che2023twinlitenet}. 
    The inference speeds of all methods are measured on the NVIDIA RTX A6000 GPU.
    }
    \label{table:novel_view}
\vspace{-3mm}
\end{table*} 
\begin{figure*}[t]

\centering
\includegraphics[width=0.977\linewidth]{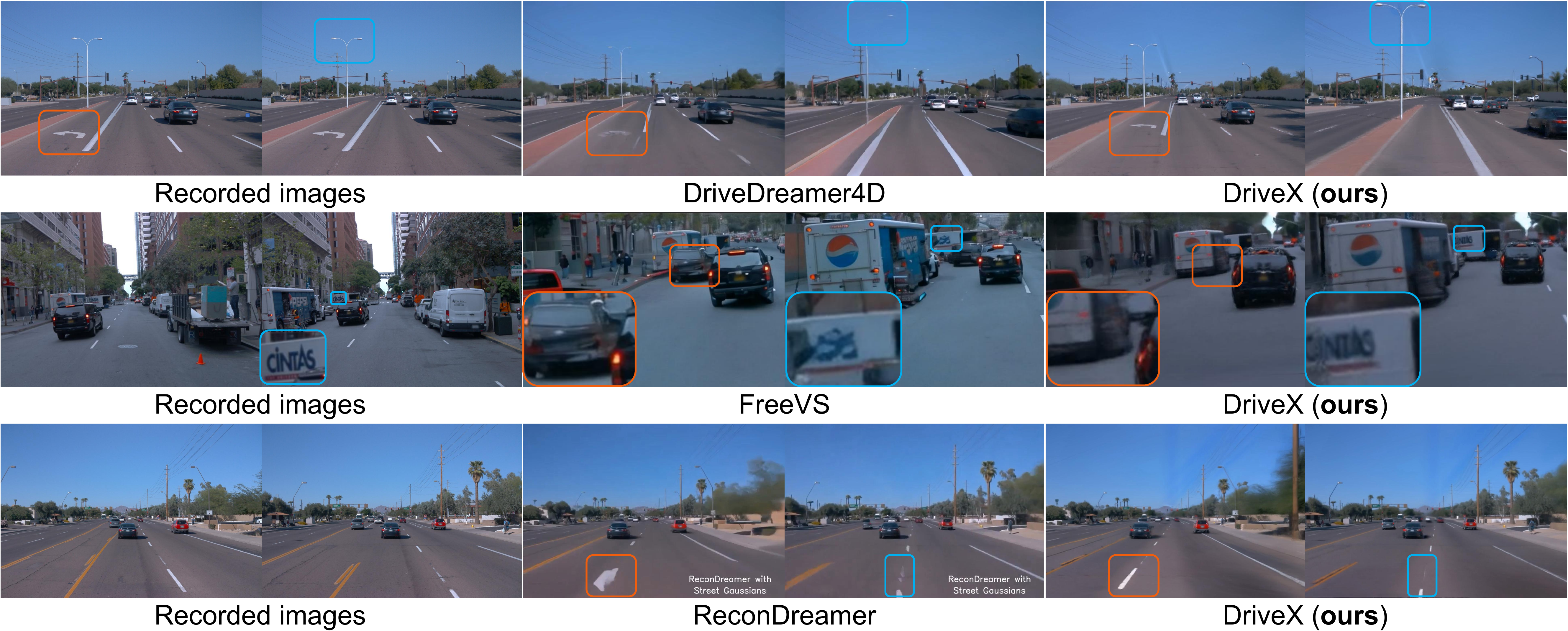}
\vspace{-3mm}
\caption{\textbf{Qualitative comparisons with methods incorperating generative model}. The results of DriveDreamer4D~\cite{zhao2024drive} and ReconDreamer~\cite{Ni2024ReconDreamer} are downloaded from their project pages. The result of FreeVS~\cite{wang2024freevs} is obtained by running its official released code. Compared to other alternatives, our method can more effectively harness the generative prior, significantly reducing hallucinations.
}
\label{fig:quanlitative_dd4d}
\vspace{-6mm}
\end{figure*}

\section{Experiments}
\label{sec:exp}

\subsection{Experimental setup}
\noindent{\bf Datasets}~~
We evaluate our framework and train the diffusion model on driving sequences from large-scale Waymo Open Dataset (WOD)~\cite{sun2020scalability}. 
Each sequence comprises surrounding videos and synchronized LiDAR point clouds capture at 10Hz. For training, we use all sequences from its training and validation set. 
The quantitative evaluation is conducted on 18 selected sequences, each containing 40 frames.
Official bounding boxes are used to crop the point cloud of dynamic traffic participants and initialize their per-frame pose. 
For fair comparison, all images are downsampled into $640 \times 960$ in both training and evaluation. More details can be found in Appendix \textcolor{iccvblue}{B}.

\noindent{\bf Metrics}~~
We evaluate the view synthesis quality on both recorded and novel trajectories. For recorded trajectory, we follow the common practice of reporting the PSNR and SSIM on these views. The test views are held out from every 5 frames.
For novel trajectory views, since the ground truth is unavailable, it is infeasible to directly assess the {\em pixel-level} fidelity. Therefore, we devised a novel benchmark to comprehensively evaluate the synthesis results at these views. 
Specifically, the Fréchet Inception Distance (FID)~\cite{heusel2017gans} between synthesized views in novel trajectories and the captured images from the original trajectories is employed to assess the realism of synthesized images at the {\em distribution level}.
Moreover, we report Lane IoU and vehicle AP (Refer to Appendix \textcolor{iccvblue}{B} for these calculations.) to evaluate the {\em distinguishability and fidelity of two safety-critical high-level traffic components}, \ie, road lanes and vehicles, in novel trajectory viewpoints. 

\noindent{\bf Implementation details}~~
For the reconstruction with calibrated camera images and LiDAR point clouds, we first initialize the scene representation~\cite{yan2024street} by optimizing on the recorded data for $T_0=50,000$ steps and then integrate generative priors for following $30,000$ steps. 
For the generation of novel trajectory videos, the initial view $P_0^{\prime}$ of a novel trajectory $\mathcal{T}^\prime$ is selected from every 3 frames, with shifting length $s$ ranging from $2m$ to $6m$. 
The generative model is initialized from ~\cite{yu2024viewcrafter}, and then fine-tuned on WOD~\cite{sun2020scalability} for $15,000$ iterations with batch size of 24 using 8 RTX A6000 GPUs. 
During the iterative refinement, the buffer for storing refined videos is updated every $K=3000$ steps.
The refine strength ({\em i.e.}~the level of noise adding to images) is set to 0.6. 
The threshold $\tau$ in~\cref{eq:mask} is set to 0.65.

\subsection{Main results}

In \cref{table:novel_view}, we quantitatively compared \ourmodel{} with several representative baselines in terms of view synthesis quality and rendering speed. 
For extrapolated novel views, we synthesized driving videos laterally shifting from the recorded trajectories by $\pm 1m$, $\pm 2m$ and $\pm 3m$.
The result demonstrates our significant
improvement on FID, AP and IoU, surpassing the leading reconstruction-based counterpart~\cite{yan2024street} with a 14.3\% decrease in FID and 56.9\% increase in IoU under the $3m$ shifting length. 
These enhancements are visually corroborated in \cref{fig:qualitative}, 
where our approach demonstrates robust rendering quality under large trajectory deviations, where safety-critical scene elements can be more clearly recognized in novel views by incorporating generative priors.
Notably, once trained, our model solely relies on the 3DGS for inference, enabling efficient rendering speed.
For views on recorded trajectories, our method also achieves comparable performance with the baseline. The slightly lower metrics often result in perceptually negligible impact on synthesized quality. 
This further indicates our method can seamlessly integrate generative prior by effectively minimizing inconsistencies of generated content.

\begin{figure}[t]\centering
\includegraphics[width=1.0\linewidth]{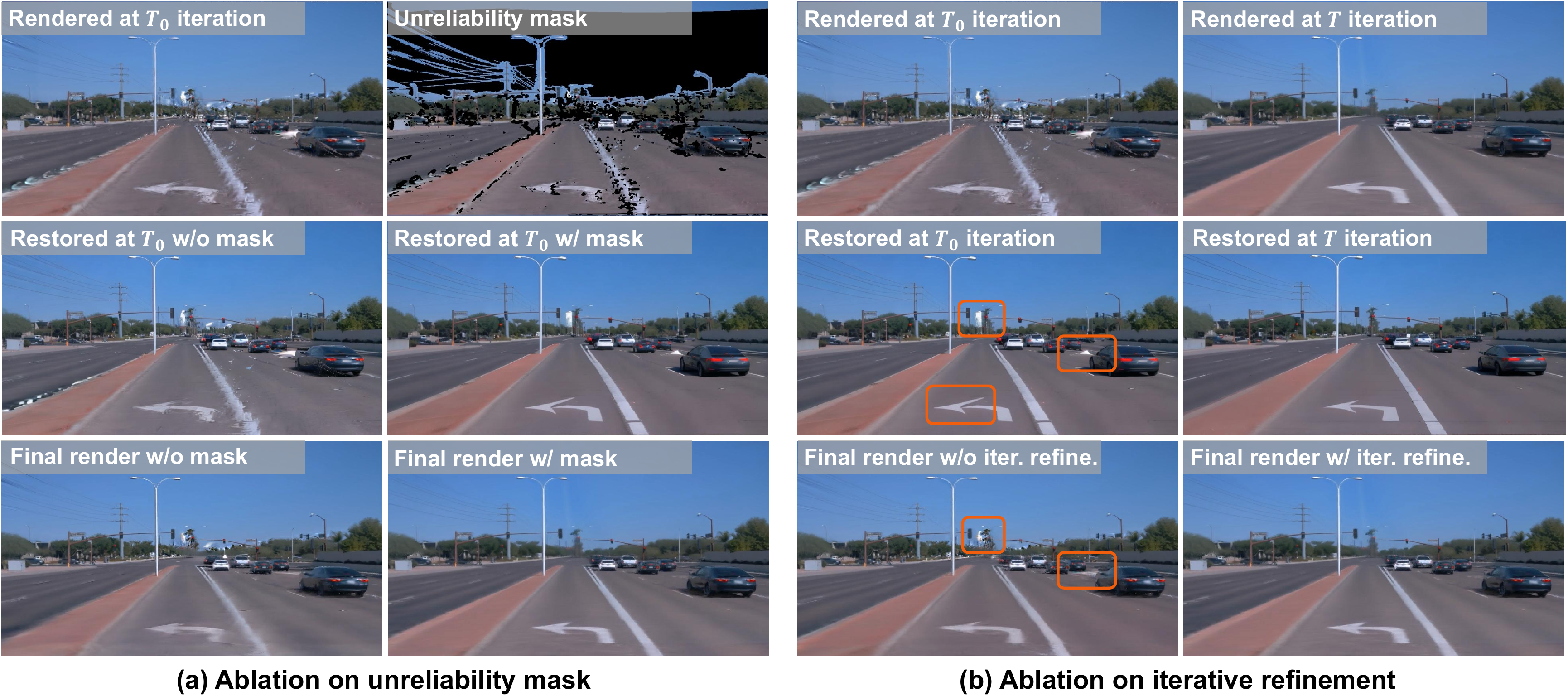}
\vspace{-7mm}
\caption{
\textbf{Ablations on (a) unreliability mask and (b) iterative refinement.}
 (a) shows the initial restored results without (middle left) and with (middle right) unreliability mask (top right).
The final rendered results exhibit significant artifacts without mask (bottom left). 
 (b) shows the rendered (top) and restored results at iteration $T_0$ (middle left) and $T$ (middle right). 
 It can be seen that single-time refinement can not completely eliminate artifacts (as shown in \textcolor[RGB]{240,94,14}{orange} boxes), 
 resulting in inferior final results compared to iterative refinement (bottom). Best viewed with zoom-in.
}
\label{fig:ablation}
\vspace{-4mm}
\end{figure} \begin{table}[t]
    \centering
    \setlength{\tabcolsep}{1.5pt}
    \footnotesize
    \begin{tabular}{l|cc|cc}
    \toprule
    & \multicolumn{2}{c|}{Lane change} &  \multicolumn{2}{c}{Lane shift $\pm 3m$} 
    \\
    & {\footnotesize NTA-IoU$\uparrow$} & {\footnotesize NTL-IoU$\uparrow$} & {\footnotesize NTA-IoU$\uparrow$} & {\footnotesize NTL-IoU$\uparrow$} \\
    \midrule
    \midrule
    DriveDreamer4D~\cite{zhao2024drive} & 0.495 & 53.42 & 0.340 & 51.32
    \\
    \midrule
    ReconDreamer~\cite{Ni2024ReconDreamer} & 0.554 & 56.63 & 0.539 & 54.58
    \\
    \midrule
    \rowcolor[gray]{.9}
     {\bf \ourmodel{} (ours)} & \bf 0.620 & \bf 58.60 & \bf 0.567 &\bf 58.29
     \\
    \toprule
    &  \multicolumn{2}{c|}{Lane shift $\pm 1m$}  &  \multicolumn{2}{c}{Lane shift $\pm 3m$} 
    \\
    & IoU$\uparrow$ & AP$\uparrow$ & IoU$\uparrow$ & AP$\uparrow$ \\
    \midrule
    \midrule
    FreeVS~\cite{wang2024freevs} &  0.2782 & 0.5777 & \bf 0.2723 & 0.5688 
    \\
    \midrule
    \rowcolor[gray]{.9}
     {\bf \ourmodel{} (ours)} & \bf 0.2880 & \bf 0.5932 & 0.2699 & \bf 0.5712
     \\
    \bottomrule
    \end{tabular}
    \vspace{-3mm}
    \caption{\textbf{Comparison with methods using generative model.} All results are averaged from 8 sequences used in ~\cite{zhao2024drive}. For \textit{lane change}, the NTA-IoU and NTL-IoU are evaluated under setting in~\cite{zhao2024drive,Ni2024ReconDreamer}, and we report the best results of~\cite{zhao2024drive,Ni2024ReconDreamer} from their original paper. 
    \textit{lane shift} is computed in the same way as our main paper. For FreeVS~\cite{wang2024freevs}, we run its official released model and measure the results on the same sequences, note that its cropped image size does not essentially impact the results.
    }
    \label{table:comp_dd4d}
\vspace{-8mm}
\end{table} 
\subsection{Comparison with methods incorporating generative prior}
In previous section, we compared our methods with popular reconstruction-based approaches as the main results because it is directly built upon these reconstruction-based solutions and primarily aims to address the view extrapolation challenge for them.
Although there are some concurrent works~\cite{zhao2024drive,Ni2024ReconDreamer,wang2024freevs,fan2024freesim,yan2024streetcrafter} sharing a similar scope with ours, most of them have not released code until our paper submission. 
Despite that, we try our best to compare with those allowing for a reasonable comparison and report the results in \cref{table:comp_dd4d}. The results indicate that our method performs better than other alternatives, which is also confirmed by qualitative results in~\cref{fig:quanlitative_dd4d}. 
Compared to DriveDreamer4D~\cite{zhao2024drive}, our approach substantially eliminates hallucinations and inconsistencies caused by insufficient constraints. 
Unlike generative-only approach FreeVS~\cite{wang2024freevs}, our method maintains continuous 3D scene representation, enabling synthesized views to preserve superior detail fidelity and better consistency across frames than those only relying on sparse LiDAR signal. 
Moreover, our method has significantly faster synthesis speed (34 FPS vs. 0.43 FPS).

\begin{table}[t]
\centering
\begin{tabular}{l|ccc}
\toprule
 & IoU$\uparrow$ & AP 
$\uparrow$ & FID$\downarrow$ \\
\midrule
\midrule
mask all & 0.2153 & 0.6334 & 102.69 \\
w/o mask & 0.2193 & 0.6341 & 76.26 \\
w/ mask  & \bf 0.2263 & \bf 0.6389 & \bf 74.34 \\
\bottomrule
\end{tabular}
\vspace{-3mm}
\caption{
\textbf{Effects of unreliability mask.} ``mask all'', ``w/o mask'', and ``w/ mask'' denotes the threshold $\tau$ in \cref{eq:mask} set to 1.0, -1.0, and 0.65, respectively.
}
\label{tab:ablation_conf_map}
\vspace{-4mm}
\end{table} \begin{table}[t]
\centering
\setlength{\tabcolsep}{4.5pt}
\begin{tabular}{l|cccc}
\toprule
Buffer interval  & IoU$\uparrow$ & AP
$\uparrow$ & FID$\downarrow$  & Time$\downarrow$  \\
\midrule
\midrule
500 & \bf 0.2271 & \bf 0.6392 & 75.33 & $4.8\times$\\
3,000 & 0.2263 & 0.6389 & \bf 74.34 & $1.6\times$\\
6,000 & 0.2228 & 0.6388 & 75.98 & $1.3\times$\\
\textit{Inf}  & 0.2194 & 0.6173 & 89.06 & $\mathbf{1.0\times}$ \\
\bottomrule
\end{tabular}
\vspace{-3mm}
\caption{\textbf{Ablations on buffer intervals.} ``\textit{Inf}'' denotes the novel trajectory supervision only generated one time at iteration $T_0$. 
}
\label{tab:ablation_iterative}
\vspace{-7mm}
\end{table} 
\subsection{Ablation study}
\label{sec:exp_ablation}

\noindent{\bf Unreliability mask}~~
The unreliability mask is designed to guide the generation of video diffusion by directly indicating where the model needs to be retained or repaired.
From \cref{tab:ablation_conf_map} and \cref{fig:ablation}(a) we can observe:
(i) When all regions are masked, the video diffusion model can only rely on sparse LiDAR projections as condition, leading to inferior results due to the inability to fully leverage information from the in-training Gaussian representation;
(ii) When the mask is disabled, the generative model can still refine the rendered results toward the learned video distribution, but usually with too conservative refinement (it also indicates that the model does not strictly follow the given mask during inpainting);
(iii) Equipped with the proposed unreliability mask, our full model achieves superior performance in both quantitative and qualitative ablation.

\noindent{\bf Iterative refinement}~~
The interval of iterative refinement offers the trade-off between quality and efficiency. 
When the novel trajectory supervision only generated once time at iteration $T_0$ (see \cref{fig:ablation}(b)), the quality of the novel view is constrained to the in-training Gaussian model, which is overfitted on the recorded trajectory. 
To the other extreme,
we are unable to afford the time cost of running the diffusion at every step.
From \cref{tab:ablation_iterative}, it can be seen that even a small interval 500 leads to $4.8\times$ training time.
Hence we finally set the interval to 3,000 to strike the balance. \section{Conclusion}
\label{sec:conclusion}

In this paper, we present a novel framework that integrates generative priors into the reconstruction of the driving scene from single-trajectory recorded videos, addressing limitations in extrapolating novel views beyond a recorded trajectory.
To reasonably leverage the video generative model to provide consistent novel view supervision, we construct a tailored inpainting-based video restoration task to refine the novel view renderings for optimizing the street scene representation.
The experiments demonstrate that the proposed method significantly improves the synthesis quality for novel trajectories. These results pave the new way for driving scene synthesis on free-form trajectories.

\section*{Acknowledgments}
This work was supported in part by National Natural Science Foundation of China (Grant No. 62376060). 
{
    \small
    \bibliographystyle{ieeenat_fullname}
    \bibliography{main}
}

\section*{Appendix}

\appendix

\section{Discussion on limitations}
\label{appendix:limits}
Beyond demonstrating the efficacy of the proposed method in synthesizing driving views along arbitrary trajectories, this paper also acknowledges several limitations. First, incorporating a diffusion-based generative model significantly increases reconstruction time. But such complexity might be unnecessary, as the required generative diversity is substantially constrained under our inpainting formulation. suggesting that a more lightweight model can be employed in the future. Second, our method currently employs a simple panning camera scheme to sample novel trajectories to be restored. However, this fix design may not be optimal for all scenarios, and leave some hand-designed components in the pipeline. Future work could consider to adaptively explore the optimal novel trajectories in optimization.

\section{More implementation details}
\label{appendix:details}

\paragraph{Metric details}
Previous driving scene reconstruction methods~\cite{yu2024sgd, han2024ggs, hwang2024vegs} typically utilize PSNR, SSIM~\citep{wang2004ssim} and LPIPS~\cite{zhang2018unreasonable} for evaluate image level similarity with ground truth. However, these metrics are inapplicable for evaluating free-form trajectories due to the absence of corresponding ground-truth images for novel trajectories. 

Therefore, inspired by~\cite{wang2024freevs,zhao2024drive,Ni2024ReconDreamer}, we propose a novel set of metrics to evaluate the recognizability of high-level traffic elements in synthesized views, based on ground truth annotations in 3D space.
For vehicles, we projected their ground truth 3D bounding boxes onto the novel views to obtain corresponding 2D bounding boxes. 
Since IoU lacks the precision granularity needed to understand performance across various intersection over union threshold.
Therefore, we calculate the Average Precision (AP[50:95]) between the projected ground truth and those detected by Faster R-CNN~\cite{ren2015faster}, which provides a more nuanced assessment than IoU.

For road lanes, NTL-IoU adopted in ~\cite{zhao2024drive,Ni2024ReconDreamer} calculates the mIoU between the lanes on synthesized views detected by~\cite{che2023twinlitenet} and the ground truth, averaging both foreground lanes and backgrounds. 
However, including the background can inflate accuracy scores due to the easy detection of non-lane areas. Therefore, following standard lane segmentation evaluation~\cite{che2023twinlitenet}, we focus specifically on IoU with only the foreground to better evaluate lane synthesis quality.

\paragraph{Warping operation $\phi$ in Eq. (\textcolor{iccvblue}{8}) of main paper}
Given a camera extrinsic matrix $E\in \mathbb{R}^{4\times4}$ and intrinsic matrix $K\in \mathbb{R}^{3\times3}$, the mapping from world points $p_{\text{world}}\in \mathbb{R}^3$ to pixel coordinates $(x,y)$ and depth $d$ can be expressed by:
\begin{align}
    \begin{bmatrix}
     p_{\text{camera}}\\1
    \end{bmatrix}
         &= (*, *, d, 1)^\top = E
    \begin{bmatrix}
     p_{\text{world}}\\1
    \end{bmatrix} 
    \\
    (x_{*}, y_{*}, d)^\top &= K p_{\text{camera}} \\
    (x, y) &= (\frac{x_{*}}{d}, \frac{y_{*}}{d})
\end{align}
Note that all the calculation above is invertible, allowing the mapping from $(x,y,d)$ to world points $p_{\text{world}}$. 

\paragraph{Validation sequences} There are 18 curated driving sequences from WOD~\cite{sun2020scalability} used for qualitative and quantitative evaluations in our paper. The names of all segments are listed below:
\begin{itemize}
    \item \verb|10359308928573410754_720_000_740_000|
    \item \verb|11450298750351730790_1431_750_1451_750|
    \item \verb|12496433400137459534_120_000_140_000|
    \item \verb|15021599536622641101_556_150_576_150|
    \item \verb|16767575238225610271_5185_000_5205_000|
    \item \verb|17860546506509760757_6040_000_6060_000|
    \item \verb|3015436519694987712_1300_000_1320_000|
    \item \verb|6637600600814023975_2235_000_2255_000|
    \item \verb|10444454289801298640_4360_000_4380_000|
    \item \verb|10588771936253546636_2300_000_2320_000|
    \item \verb|10625026498155904401_200_000_220_000|
    \item \verb|11017034898130016754_697_830_717_830|
    \item \verb|1191788760630624072_3880_000_3900_000|
    \item \verb|11928449532664718059_1200_000_1220_000|
    \item \verb|14810689888487451189_720_000_740_000|
    \item \verb|4414235478445376689_2020_000_2040_000|
    \item \verb|6242822583398487496_73_000_93_000|
    \item \verb|7670103006580549715_360_000_380_000|
\end{itemize}

Unless stated otherwise, all quantitative ablation experiments are conducted on segments \verb|15021599536622641101_556_150_576_150| and \verb|3015436519694987712_1300_000_1320_000|, while the results are evaluated on novel trajectories with lateral shift $3m$.

\section{Additional ablations}
\begin{table}[t]
\centering
\begin{tabular}{l|ccccc}
\toprule
Noise level & IoU$\uparrow$ & AP 
$\uparrow$ & FID$\downarrow$ & Time$\downarrow$ \\
\midrule
\midrule
0.4  & 0.2184 & 0.6202 & 96.21 &  $\mathbf{1.0\times}$\\
0.6  & \bf 0.2263 & \bf 0.6389  & \bf 74.34 & $1.1\times$\\
0.8  & 0.2192  & 0.6259 & 81.55 &  $1.2\times$\\
1.0  & 0.2113  & 0.6293 & 106.01 & $1.3\times$\\
\bottomrule
\end{tabular}
\caption{\textbf{Effect of the noise level.} The noise level $l$ denotes adding the noise corresponding to the timestep $t=\lfloor lN \rfloor$, where $N$ is the total step needed for denoising from scratch.} 
\label{tab:ablation_noise}
\end{table} \begin{table}[t]
\centering
\setlength{\tabcolsep}{4pt}
\begin{tabular}{l|c|ccc}
\toprule
Gaussian model & w/ ours & IoU$\uparrow$ & AP 
$\uparrow$ & FID$\downarrow$ \\
\midrule
\midrule
PVG~\cite{chen2023periodic} & & 0.0200 & 0.5401 & 118.41  \\
PVG~\cite{chen2023periodic} & \checkmark & \bf 0.1872 & \bf 0.6197 & \bf 91.85 \\
\midrule
StreetGaussian~\cite{yan2024street} & & 0.1217 & 0.6124 & 103.08 \\
StreetGaussian~\cite{yan2024street} & \checkmark & \bf 0.2263 & \bf 0.6389 & \bf 74.34 \\
\bottomrule
\end{tabular}
\caption{\textbf{Ablation study on different Gaussian model.}}
\label{tab:ablation_representation}
\end{table} \begin{table}[t]
\centering
\begin{tabular}{l|cccc}
\toprule
 &  IoU$\uparrow$ & AP 
$\uparrow$ & FID$\downarrow$  \\
\midrule
\midrule
w/o progress. & 0.2204 & 0.6372 & 75.41 \\
w/o warp & 0.2196 & 0.6357 & 105.77 \\
Full model & \bf 0.2263 & \bf 0.6389 & \bf 74.34 \\

\bottomrule
\end{tabular}
\caption{\textbf{Ablations on other design choices.} ``w/o progress'' denotes that the shift length is fixed at $4m$ throughout the optimization. ``w/o warp'' means that the denoising is started from the noisy rendered images rather than warped views.}
\label{tab:ablation_design}
\end{table} 
\paragraph{Refine strength}
When we refine the rendered videos using video diffusion, we also study the effect the refine strength (level of noise added to the images). In \cref{tab:ablation_noise}, we show the metrics for novel trajectory synthesis (IoU, AP, and FID), along with the total optimization time.
With a low refine strength, the refined image adheres to the artifact-heavy rendered image, resulting in minimal improvement. On the other hand, a high refine strength yields better novel views but may not keep fidelity to the original scene. Additionally, higher refine strength requires more denoising steps, increasing the training time. Consequently, refine strength of 0.6 is found to be the best balance.

\paragraph{Different Gaussian model} 
Although we adopt static 3D Gaussian splatting~\cite{yan2024street} as the default scene representation in previous experiments, our method is also capable of synergy with different choices.
To demonstrate the versatility of \ourmodel{}, we also evaluate our method on a representative dynamic Gaussian representation for driving view synthesis~\cite{chen2023periodic}.
From \cref{tab:ablation_representation}, 
we can observe that our strategy consistently reduces the FID by a large margin by substantially reducing artifacts in novel views, 
and effectively recovering street lanes and cars ({\em i.e.}~IoU, AP) that are entirely indiscernible in the baseline.

\begin{figure}[ht]
\centering
\includegraphics[width=0.98\linewidth]{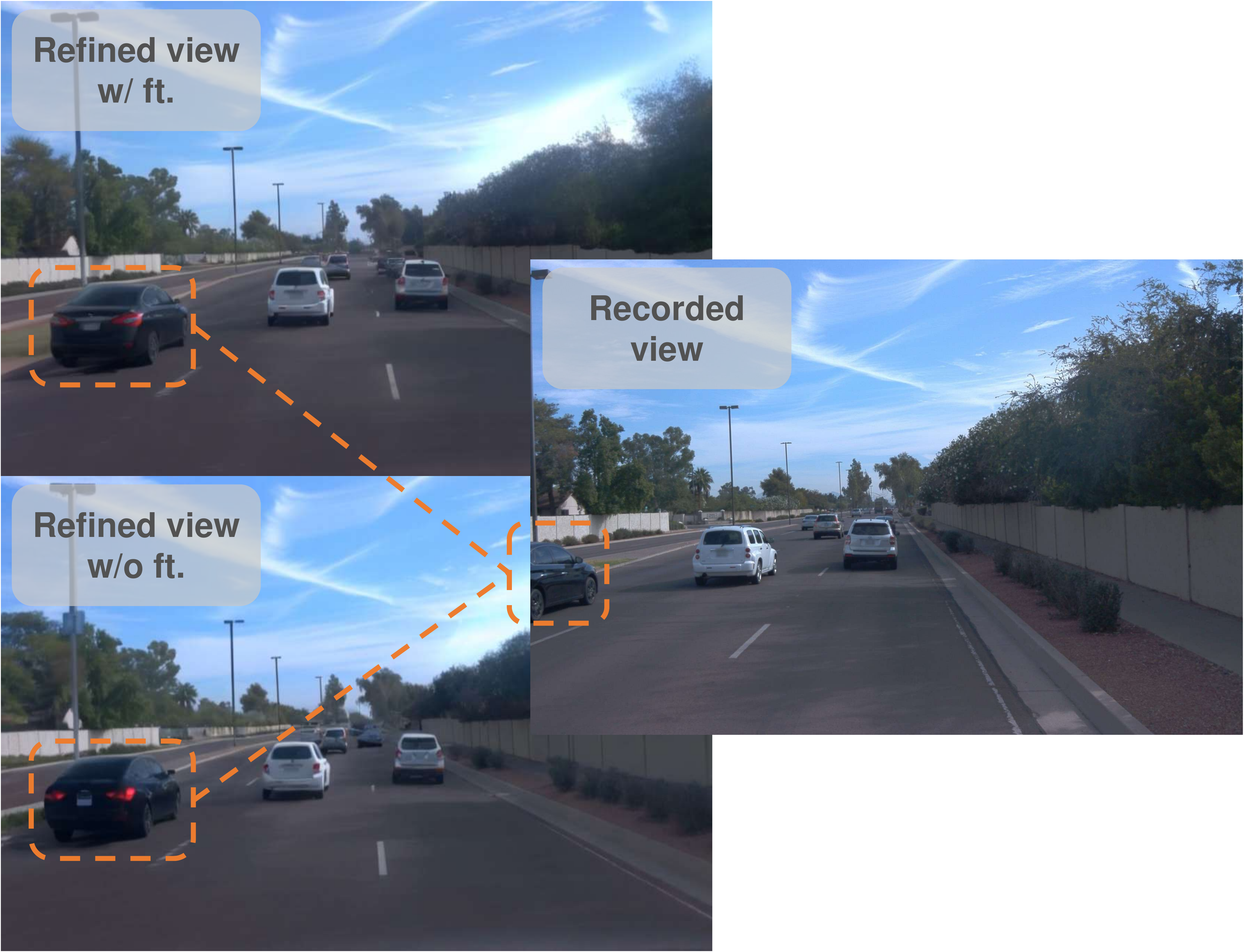}
\vspace{-2mm}
\caption{
\textbf{Domain-specific fine-tuning enhances the style adaptation on driving videos.} Here the appearance of the black car refined by model without fine-tuning seems to lack details and has unrealistic material. 
}
\label{fig:ablation_diffusion_prior}
\vspace{-3mm}
\end{figure} 
\noindent{\bf Projected LiDAR points as auxiliary oracle}~~
Although using the masked rendered image as the only condition is sufficient to achieve satisfactory performance in most scenes, incorporating projected LiDAR points as the additional oracle can greatly improve robustness in some difficult cases. In these scenes, initial poor novel view renderings would result in large masked regions, LiDAR projection can provide direct guidance for precisely restoring them. This resulted enhancement is reflected by the quantitative comparison in \cref{tab:ablation_diffusion_prior}.

\begin{figure*}[t]
    \centering 
    \setlength{\tabcolsep}{0.5pt}
    \renewcommand{\arraystretch}{0.15}
    \begin{tabular}{cccccc} 
     \rotatebox{90}{\small{w/o ours}} 
     & 
     \raisebox{-0.2\height}{\includegraphics[width=.19\textwidth,clip]{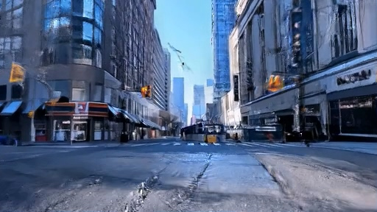}}
    &
    \raisebox{-0.2\height}{\includegraphics[width=.19\textwidth,clip]{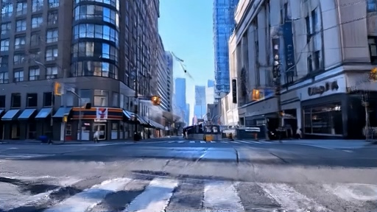}}
    &
    \raisebox{-0.2\height}{\includegraphics[width=.19\textwidth,clip]{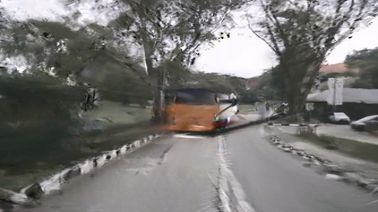}}
    &
    \raisebox{-0.2\height}{\includegraphics[width=.19\textwidth,clip]{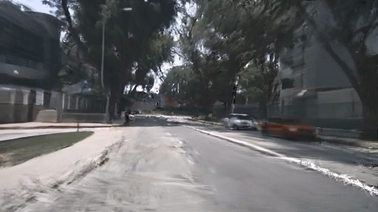}}
    &
    \raisebox{-0.2\height}{\includegraphics[width=.19\textwidth,clip]{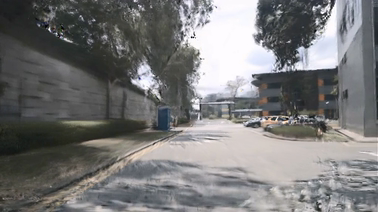}}
 \\
 \\
    \rotatebox{90}{\small{w/ ours}} 
    & 
    \raisebox{-0.2\height}{\includegraphics[width=.19\textwidth,clip]{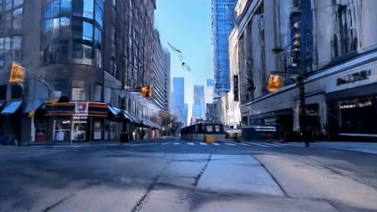}}
    &
    \raisebox{-0.2\height}{\includegraphics[width=.19\textwidth,clip]{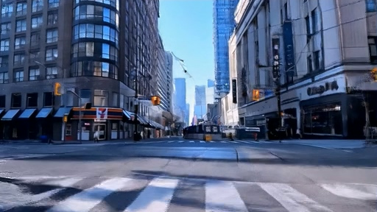}}
    &
    \raisebox{-0.2\height}{\includegraphics[width=.19\textwidth,clip]{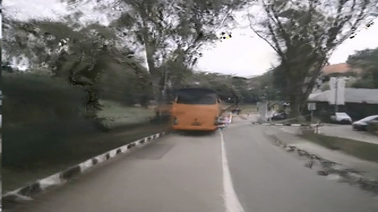}}
    &
    \raisebox{-0.2\height}{\includegraphics[width=.19\textwidth,clip]{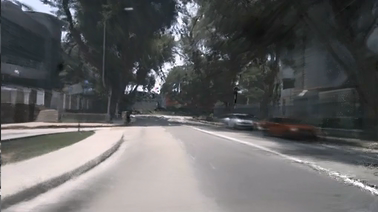}}
    &
    \raisebox{-0.2\height}{\includegraphics[width=.19\textwidth,clip]{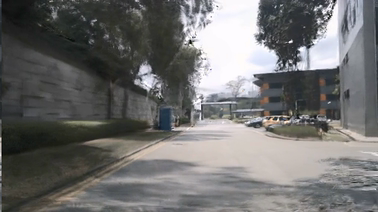}}
 \\

    \end{tabular}
\caption{\textbf{
    Reconstruction results on generated videos.} All ground truth videos used for reconstruction are generated by Vista~\cite{gao2024vista}. The first and the second rows show the novel trajectory synthesis results by the baseline~\cite{yan2024street} and our method, respectively. }
\label{fig:qualitative_gen}
\end{figure*} \begin{table}[t]
\centering
\begin{tabular}{ccc|ccc}
\toprule
w/ & w/ & w/  & \multirow{2}*{IoU$\uparrow$} & \multirow{2}*{AP 
$\uparrow$} & \multirow{2}*{FID$\downarrow$} \\
render & LiDAR & ft. &  & &  \\
\midrule
\midrule
\checkmark & & & 0.2092 & 0.6257 & 90.30 \\
\checkmark & \checkmark & & 0.2123 & 0.6334 & 86.30 \\
 & \checkmark & \checkmark & 0.2153 & 0.6225 & 75.18 \\
\checkmark & \checkmark & \checkmark & \bf 0.2263 & \bf 0.6389 & \bf 74.34 \\ 
\bottomrule
\end{tabular}
\vspace{-1mm}
\caption{
\textbf{Ablations on the components of generative model.} ``w/ render'' means using masked novel view image as condition. ``w/ LiDAR'' denotes adopting LiDAR projection as condition. ``w/ ft.'' represents training diffusion model on driving video data~\cite{sun2020scalability}. 
}
\label{tab:ablation_diffusion_prior}
\vspace{-6mm}
\end{table} 
\noindent{\bf Fine-tuning generative model on driving videos}~~
The comparison between the $2^{nd}$ and $4^{th}$ row of  \cref{tab:ablation_diffusion_prior} demonstrates that the diffusion model fine-tuned on in-domain driving videos outperforms that trained solely on general-purpose 3D datasets~\cite{yu2024viewcrafter}, especially in FID. This improvement comes from the generative model's adaptation to the style of real driving scenes. As shown in \cref{fig:ablation_diffusion_prior}, the counterpart without domain-specific fine-tuning tends to generate cartooned foreground vehicles without detailed texture. Therefore, beyond the improvement in quantitative metrics, it also simplifies our pipeline by removing the need for some hand-designed processing, such as lowering the refine weight for foreground vehicles.

\paragraph{Other designs}
We also ablate other design choices in~\cref{tab:ablation_design}, including progressive increasing shifted length and warped condition. The results show that the final quality degrades without any element, which indicates their helpfulness on the rendering quality and the stability of optimization.

\begin{table}[t]
    \centering
\setlength{\tabcolsep}{8pt}
    \begin{tabular}{l|ccc}
    \toprule
        \multirow{2}{*}{Method} & \multicolumn{3}{c}{FID $\downarrow$} \\
        & $\pm 0 m$ & $\pm 1 m$ & $\pm 2 m$ \\
        \midrule
        \midrule
        Recon-only~\cite{yan2024street}
        &  45.45 & 96.76 & 146.61 \\
        \bf \ourmodel{} (ours)   
        &  \bf 45.44 & \bf 92.77 & \bf 142.73  \\
        \bottomrule
    \end{tabular}
\caption{\textbf{Quantitative comparison on generated video.} 
    We compared the FID score on videos generated by Vista~\cite{gao2024vista}. 
    }
    \label{table:generation_gt}

\end{table}

\section{Application on generated videos}
Another intriguing application of the proposed method is transforming AI-generated video into a re-enactable driving world. However, the AI-generated videos present more severe challenges because of the lack of corresponding LiDAR depth and accurate pixel-to-pixel matching between different frames. 
These challenges further necessitate the novel view supervision provided by the generative priors. Therefore, we devise an additional recipe for the robust optimization of the AI-generated videos.

\paragraph{Details} Specifically, since the generated videos cannot be precisely aligned with the conditioned camera parameters, we employ a feed-forward approach~\cite{zhang2024monst3r} to estimate the camera parameters and corresponding depths for each frame. For scenes containing moving vehicles, we manually annotate their tracklets as the initialization for the local coordinate frames.
During optimization, we fine-tune the camera pose to mitigate potential prediction errors. 
In the computation of unreliability, as the generated video is monocular, we use the nearest previous frames that can include the current shifted view content as the auxiliary source images in Eq. (\textcolor{iccvblue}{9}) of main paper to ensure well-defined unreliability scores for all areas of the novel view. Besides, we include an entropy loss for the opacity of each Gaussian to prevent potential overfitting. 

\paragraph{Results} To quantitatively evaluate this capability, 
we compared \ourmodel{} and the reconstruction-only baseline~\cite{yan2024street} on 12 driving videos generated by Vista~\cite{gao2024vista}. Due to the evident stylistic gap between the generated videos and those in WOD, we only use the diffusion model trained on general-purpose 3D datasets as generative prior without LiDAR projection as condition.
The FID of videos rendered across different novel trajectories is reported in \cref{table:generation_gt}. The results indicate that the realism of rendered videos is effectively preserved within a certain range of shifted length. 
Compared to the reconstruction-only baseline, our \ourmodel{} achieves 4.1\% and 2.6\% FID reduction in shifting length of $\pm 1m$ and $\pm 2m$ respectively. To obtain real-world distance, we align the predicted depth and LiDAR depth of the first frame by the least square estimation. 
These advancements are also clearly illustrated in \cref{fig:qualitative_gen}. As can be seen, the baseline method exhibits significant degradation in views that are far from the recorded trajectory. 

\begin{figure}[t]
\centering
\includegraphics[width=1.\linewidth]{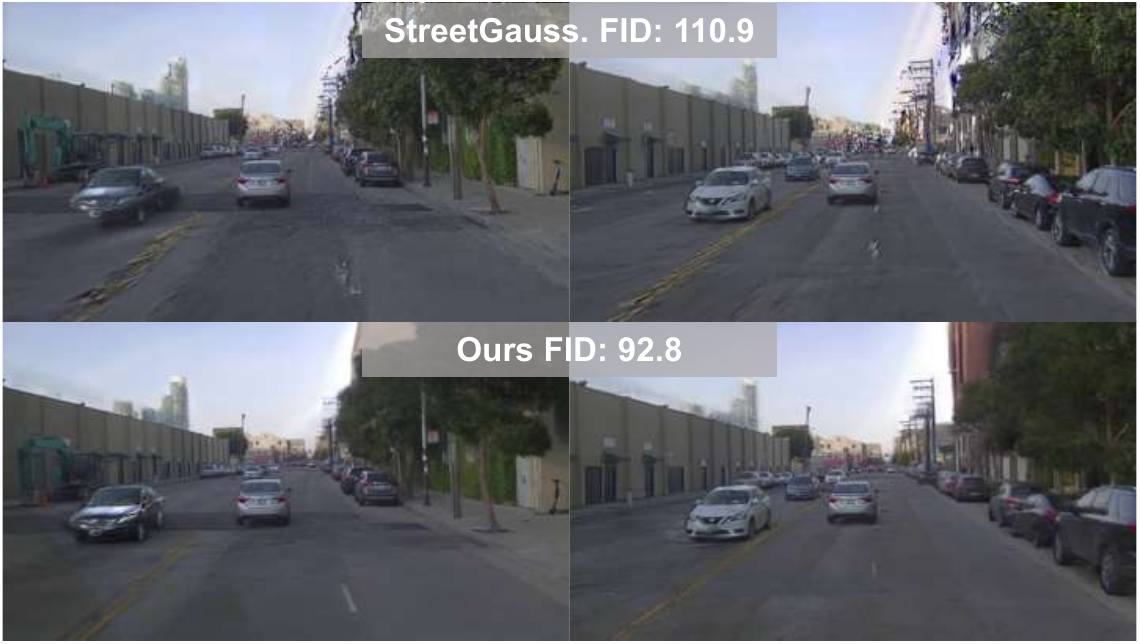}
\caption{
\textbf{Results on the PandaSet dataset~\cite{xiao2021pandaset}.} The results are on the novel trajectory shifted $2.0m$ from the recoded trajectory.}
\label{fig:results_pandaset}
\end{figure} 
\section{Results on the PandaSet dataset}
To further validate the versatility of the proposed method on a broader range of scene types and sensor configurations, we apply our approach to two sequences indexed $001$ and $003$ from the PandaSet dataset~\cite{xiao2021pandaset}, each consisting of 80 frames. The image resolution is downsampled to $450 \times 800$ for both training and evaluation. 
In~\cref{fig:results_pandaset}, we compared our method with the state-of-the-art reconstruction baseline StreetGaussian~\cite{yan2024street} on the trajectory with shift length $2.0m$. The result suggests that our method can significantly reduce artifacts in synthesized images and improve the discernibility of safety-critical traffic elements, which is also corroborated by our better FID.

\section{More quanlitative results}
\begin{figure}[t]
\centering
\includegraphics[width=1.\linewidth]{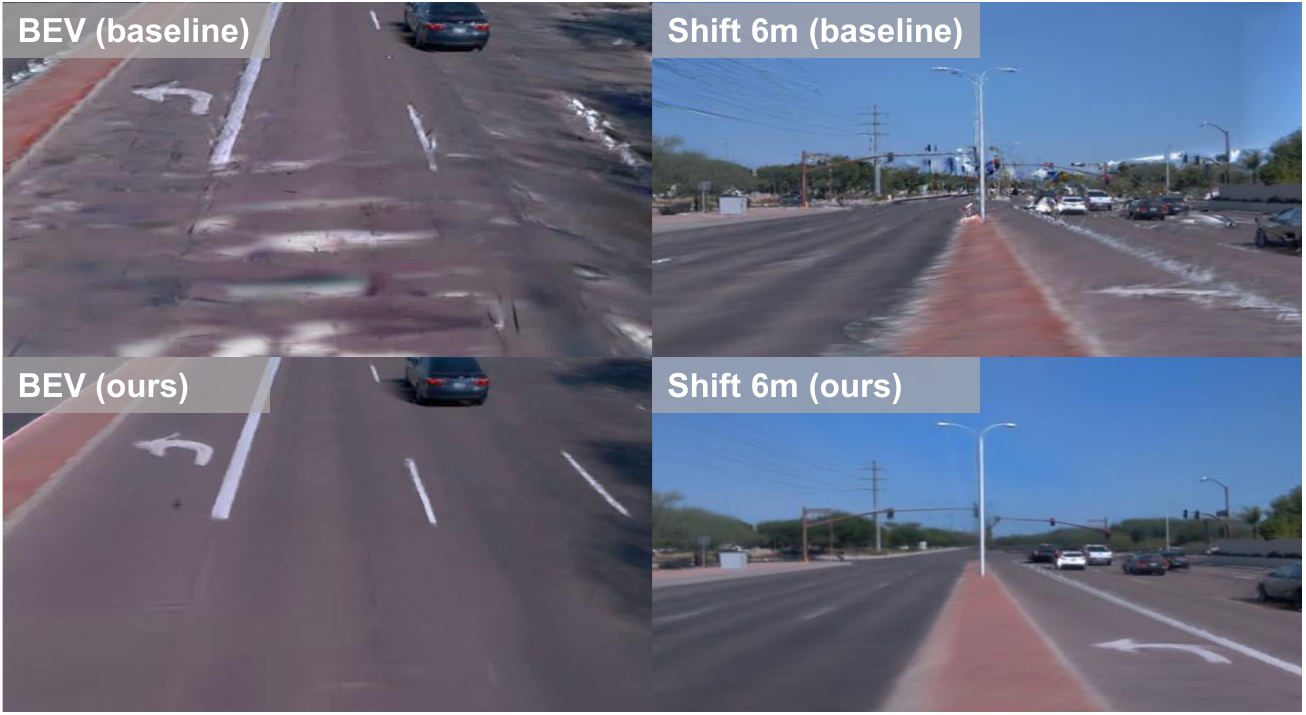}
\caption{
\textbf{Results on more aggressive view change.} 
} 
\label{fig:qualitative_challange}
\end{figure} \begin{figure}[t]
\centering
\includegraphics[width=0.977\linewidth]{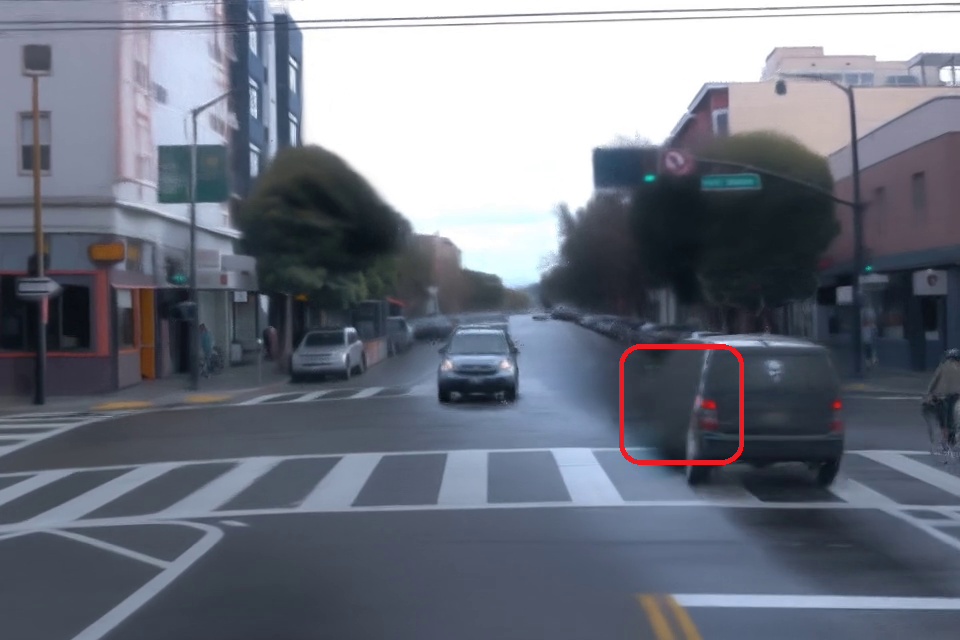}
\caption{
\textbf{Failure cases.} Our method cannot \textit{reconstruct} a feasible appearance from scratch for the side of a vehicle that is invisible from all recorded views.
}
\label{fig:failure_case}
\end{figure} \paragraph{More aggressive view change.} 
We test the proposed method on some challenging cases to assess its robustness. 
In~\cref{fig:qualitative_challange} (left), the view is rendered from the camera elevated by $3m$ above the recorded trajectory, with a pitch angle of $30^\circ$. In~\cref{fig:qualitative_challange} (right), we increase the lateral shift length to $6m$.
To accommodate these aggressively changed views, we include them in the refined trajectory during optimization.
As shown in the results, the baseline method almost totally fails to synthesize meaningful images under these viewpoints, while our solution substantially improves the fidelity.

\paragraph{Failure case.} Despite the robustness and generalization ability on diverse scenes as demonstrated above, it should be noted that the proposed framework only aims to provide regularization for unconstrained reconstruction.
As a reconstruction pipeline, it does not work well on totally unseen regions of the scene. 
For example, in~\cref{fig:failure_case}, our method cannot create a feasible appearance from scratch for the side of a vehicle that is invisible from all recorded views and thus lacks proper geometry initialization before the refinement stage.

\begin{figure}[t]
\centering
\includegraphics[width=1.\linewidth]{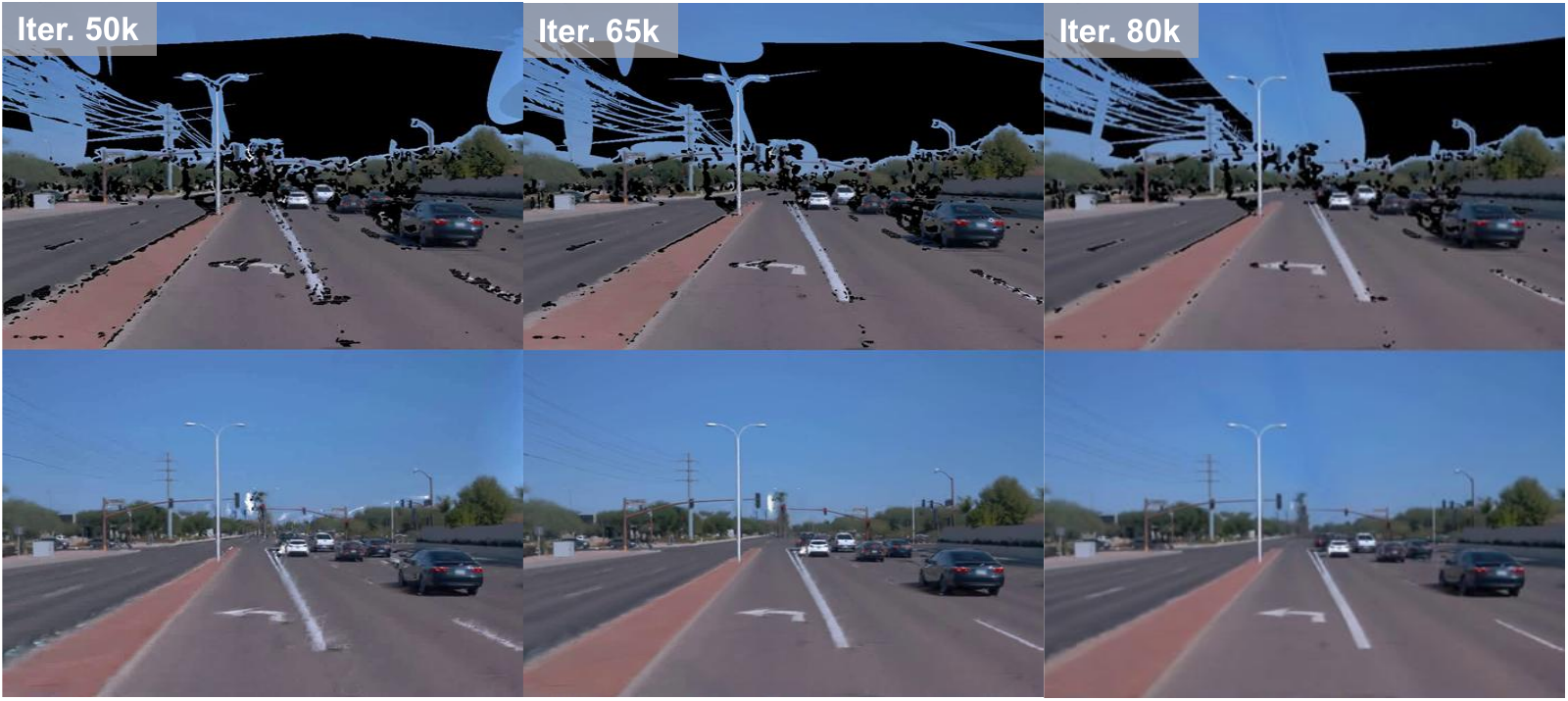}
\caption{
\textbf{The evolution of unreliability mask during optimization.} 
} 
\label{fig:qualitative_evolution}
\end{figure} \paragraph{The evolution of unreliability mask during optimization.}
In~\cref{fig:qualitative_evolution}, we present novel view renderings and the corresponding unreliability masks at different training iterations. 
The results clearly show that the reliable regions gradually expand during iterative refinement, demonstrating the effectiveness of the proposed unreliability mask mechanism--it can serve as a reliable cue for the rendered images rather than introducing compounding errors. 
Additionally, the distribution of the mask indicates that unreliable regions mainly appear near edge areas of the visual elements, which is the rationale behind the adoption of edge-aware inpainting.

\textbf{More comparisons with sota methods} The video comparison with the state-of-the-art alternates can be found in our \href{https://fudan-zvg.github.io/DriveX}{project page}. 
\end{document}